\documentclass[11pt,a4paper]{article}
\usepackage[hyperref]{emnlp2018}
\usepackage{times}
\usepackage{latexsym}

\usepackage{algorithm}
\usepackage{algorithmicx}
\usepackage{algpseudocode}
\usepackage{url}
\usepackage{amsmath}
\usepackage{amssymb}
\usepackage{graphicx}
\usepackage{multirow}

\aclfinalcopy 
\def\aclpaperid{1536} 

\title{Semi-Autoregressive Neural Machine Translation}

\author{Chunqi Wang\thanks{\ \ Part of this work was done when the author was at Institute of Automation, Chinese Academy of Sciences.}\qquad Ji Zhang\qquad Haiqing Chen \\
	Alibaba Group \\
	{\tt \{shiyuan.wcq, zj122146, haiqing.chenhq\}@alibaba-inc.com}}

\date{}

\begin{document}
\maketitle
\begin{abstract}
Existing approaches to neural machine translation are typically autoregressive models.
While these models attain state-of-the-art translation quality, they are suffering from low parallelizability and thus slow at decoding long sequences.
In this paper, we propose a novel model for fast sequence generation --- the semi-autoregressive Transformer (SAT).
The SAT keeps the autoregressive property in global but relieves in local and thus is able to produce multiple successive words in parallel at each time step.
Experiments conducted on English-German and Chinese-English translation tasks show that the SAT achieves a good balance between translation quality and decoding speed.
On WMT'14 English-German translation, the SAT achieves 5.58$\times$ speedup while maintains 88\% translation quality, significantly better than the previous non-autoregressive methods. When produces two words at each time step, the SAT is almost lossless (only 1\% degeneration in BLEU score).
\end{abstract}

\section{Introduction}
Neural networks have been successfully applied to a variety of tasks, including machine translation. The encoder-decoder architecture is the central idea of neural machine translation (NMT). The encoder first encodes a source-side sentence ${\bf x}= x_1\dots x_m$ into hidden states and then the decoder generates the target-side sentence ${\bf y} = y_1\dots y_n$ from the hidden states according to an autoregressive model
$$p(y_t|y_1\dots y_{t-1},{\bf x})$$
Recurrent neural networks (RNNs) are inherently good at processing sequential data. \newcite{sutskever2014sequence,cho2014learning} successfully applied RNNs to machine translation. \newcite{bahdanau2014neural} introduced attention mechanism into the encoder-decoder architecture and greatly improved NMT. GNMT \cite{Wu2016Google} further improved NMT by a bunch of tricks including residual connection and reinforcement learning.

\begin{figure}
	\centering
	\includegraphics[width=\linewidth]{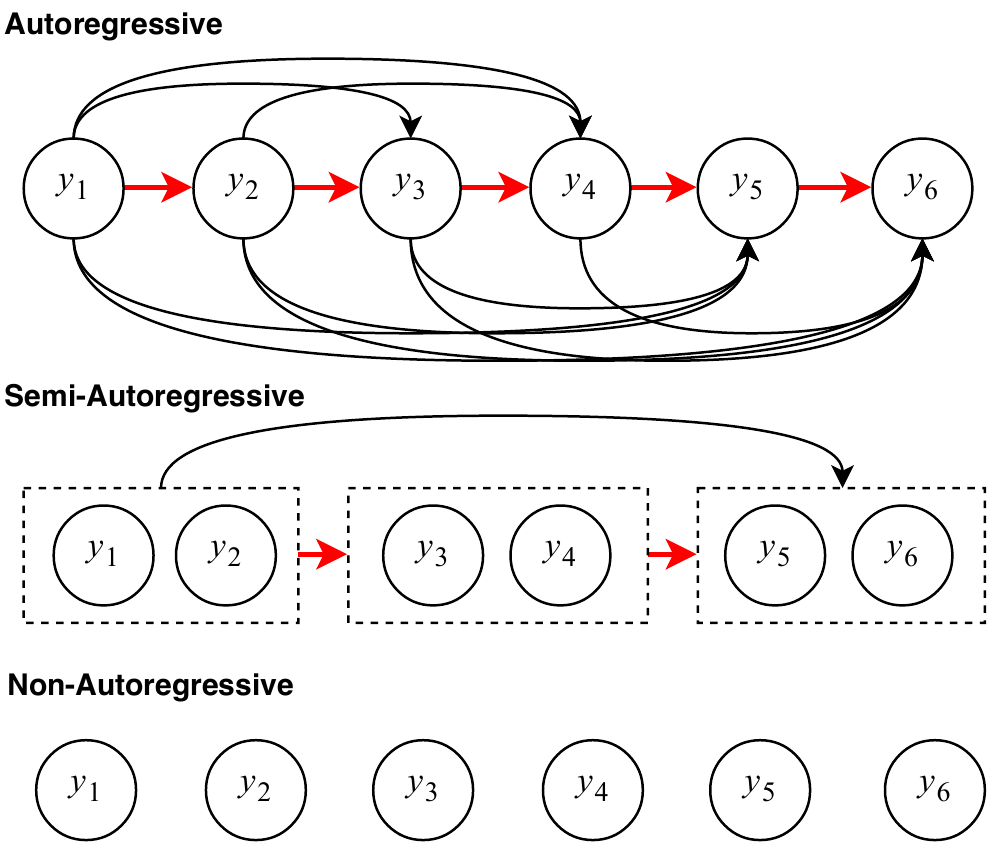}
	\caption{\label{fig:auto}The different levels of autoregressive properties. Lines with arrow indicate dependencies. We mark the longest dependency path with bold red lines. The length of the longest dependency path decreases as we relieve the autoregressive property. An extreme case is \emph{non-autoregressive}, where there is no dependency at all.}
\end{figure}

The sequential property of RNNs leads to its wide application in language processing. However, the property also hinders its parallelizability thus RNNs are slow to execute on modern hardware optimized for parallel execution. As a result, a number of more parallelizable sequence models were proposed such as ConvS2S \cite{Gehring2017Convolutional} and the Transformer \cite{vaswani2017attention}. These models avoid the dependencies between different positions in each layer thus can be trained much faster than RNN based models. When inference, however, these models are still slow because of the autoregressive property.

A recent work \cite{gu2017non} proposed a non-autoregressive NMT model that generates all target-side words in parallel. While the parallelizability is greatly improved, the translation quality encounter much decrease. 
In this paper, we propose the semi-autoregressive Transformer (SAT) for faster sequence generation.
Unlike \newcite{gu2017non}, the SAT is semi-autoregressive, which means it keeps the autoregressive property in global but relieves in local. 
As the result, the SAT can produce multiple successive words in parallel at each time step.
Figure \ref{fig:auto} gives an illustration of the different levels of autoregressive properties.

Experiments conducted on English-German and Chinese-English translation show that compared with non-autoregressive methods, the SAT achieves a better balance between translation quality and decoding speed. On WMT'14 English-German translation, the proposed SAT is 5.58$\times$ faster than the Transformer while maintaining 88\% of translation quality. Besides, when produces two words at each time step, the SAT is almost lossless.

It is worth noting that although we apply the SAT to machine translation, it is not designed specifically for translation as \newcite{gu2017non,lee2018deterministic}. The SAT can also be applied to any other sequence generation task, such as summary generation and image caption generation.

\section{Related Work}
Almost all state-of-the-art NMT models are autoregressive \cite{sutskever2014sequence,bahdanau2014neural,Wu2016Google,Gehring2017Convolutional,vaswani2017attention}, meaning that the model generates words one by one and is not friendly to modern hardware optimized for parallel execution.
A recent work \cite{gu2017non} attempts to accelerate generation by introducing a non-autoregressive model.
Based on the Transformer \cite{vaswani2017attention}, they made lots of modifications.
The most significant modification is that they avoid feeding the previously generated target words to the decoder, but instead feeding the source words, to predict the next target word.
They also introduced a set of latent variables to model the \emph{fertilities} of source words to tackle the multimodality problem in translation.
\newcite{lee2018deterministic} proposed another non-autoregressive sequence model based on iterative refinement. The model can be viewed as both a latent variable model and a conditional denoising autoencoder. They also proposed a learning algorithm that is hybrid of lower-bound maximization and reconstruction error minimization.

The most relevant to our proposed semi-autoregressive model is \cite{kaiser2018fast}. They first autoencode the target sequence into a shorter sequence of discrete latent variables, which at inference time is generated autoregressively, and finally decode the output sequence from this shorter latent sequence in parallel. What we have in common with their idea is that we have not entirely abandoned autoregressive, but rather shortened the autoregressive path.

A related study on realistic speech synthesis is the parallel WaveNet \cite{oord2017parallel}. The paper introduced \emph{probability density distillation}, a new method for training a parallel feed-forward network from a trained WaveNet \cite{van2016wavenet} with no significant difference in quality.

There are also some work share a somehow simillar idea with our work: character-level NMT \cite{chung2016character,lee2016fully} and chunk-based NMT \cite{zhou2017chunk,ishiwatari2017chunk}. Unlike the SAT, these models are not able to produce multiple tokens (characters or words) each time step. \newcite{oda2017neural} proposed a bit-level decoder, where a word is represented by a binary code and each bit of the code can be predicted in parallel.

\section{The Transformer}
Since our proposed model is built upon the Transformer \cite{vaswani2017attention}, we will briefly introduce the Transformer.
The Transformer uses an encoder-decoder architecture. We describe the encoder and decoder below.

\subsection{The Encoder}
From the source tokens, learned embeddings of dimension $d_{model}$ are generated which are then modified by an additive positional encoding.
The positional encoding is necessary since the network does not leverage the order of the sequence by recurrence or convolution. The authors use additive encoding which is defined as:
\begin{align*}
PE(pos, 2i) &= sin(pos/10000^{2i/d_{model}})\\
PE(pos, 2i+1) &= cos(pos/10000^{2i/d_{model}})
\end{align*}
where $pos$ is the position of a word in the sentence and $i$ is the dimension. The authors chose this function because they hypothesized it would allow the model to learn to attend by relative positions easily.
The encoded word embeddings are then used as input to the encoder which consists of $N$ blocks each containing two layers: (1) a multi-head attention layer, and (2) a position-wise feed-forward layer.

Multi-head attention builds upon scaled dot-product attention, which operates on a query Q, key K and value V:
$$Attention(Q,K,V) = softmax(\frac{QK^T}{\sqrt{d_k}})V$$
where $d_k$ is the dimension of the key. The authors scale the dot product by $1/\sqrt{d_k}$ to avoid the inputs to softmax function growing too large in magnitude. 
Multi-head attention computes $h$ different queries, keys and values with $h$ linear projections, computes scaled dot-product attention for each query, key and value, concatenates the results, and projects the concatenation with another linear projection:
\begin{align*}
&H_i = Attention(QW_i^Q, KW_i^K, VW_i^V) \\
&MultiHead(Q,K,V) = Concat(H_1,\dots H_h) 
\end{align*}
in which $W_i^Q, W_i^K \in \mathbb{R}^{d_{model}\times d_k}$ and $W_i^V \in \mathbb{R}^{d_{model}\times d_v}$. The attention mechanism in the encoder performs attention over itself ($Q=K=V$), so it is also called self-attention.

The second component in each encoder block is a position-wise feed-forward layer defined as:
$$FFN(x) = max(0, xW_1+b_1)W_2+b_2$$
where $W_1\in \mathbb{R}^{d_{model}\times d_{ff}}$, $W_2\in \mathbb{R}^{d_{ff}\times d_{model}}$, $b_1\in \mathbb{R}^{d_{ff}}$, $b_2\in\mathbb{R}^{d_{model}}$.

For more stable and faster convergence, residual connection \cite{he2016deep} is applied to each layer, followed by layer normalization \cite{ba2016layer}.
For regularization, dropout \cite{srivastava2014dropout} are applied before residual connections.

\begin{figure}[!htb]
	\centering
	\includegraphics[width=\linewidth]{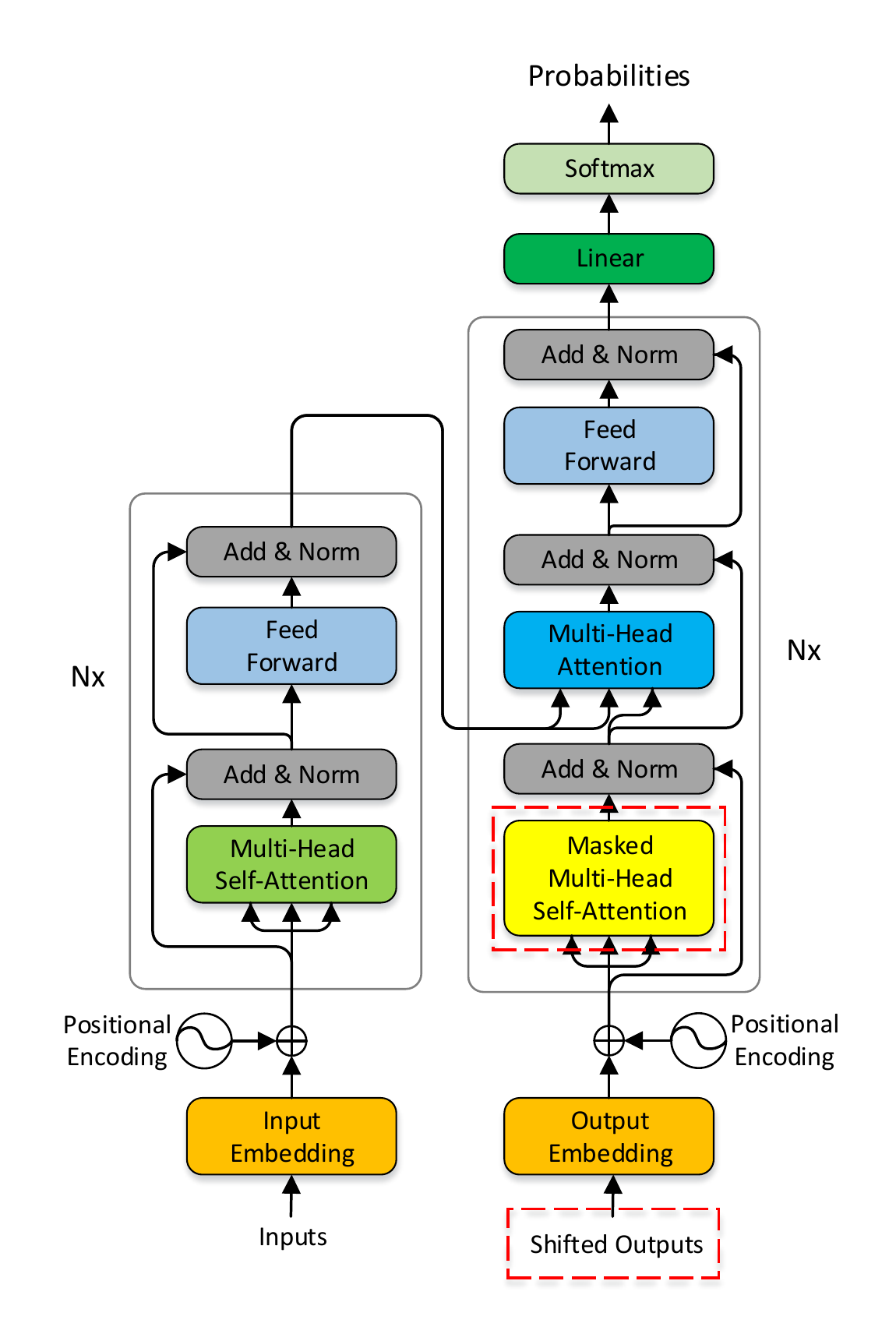}
	\caption{\label{fig:transformer}The architecture of the Transformer, also of the SAT, where the red dashed boxes point out the different parts of these two models.}
\end{figure}

\subsection{The Decoder}
The decoder is similar with the encoder and is also composed by $N$ blocks. In addition to the two layers in each encoder block, the decoder inserts a third layer, which performs multi-head attention over the output of the encoder.

It is worth noting that, different from the encoder, the self-attention layer in the decoder must be masked with a causal mask, which is a lower triangular matrix, to ensure that the prediction for position $i$ can depend only on the known outputs at positions less than $i$ during training.

\section{The Semi-Autoregressive Transformer}
We propose a novel NMT model---the Semi-Autoregressive Transformer (SAT)---that can produce multiple successive words in parallel.
As shown in Figure \ref{fig:transformer}, the architecture of the SAT is almost the same as the Transformer, except some modifications in the decoder.

\subsection{Group-Level Chain Rule}
Standard NMT models usually factorize the joint probability of a word sequence $y_1\dots y_n$ according to the word-level chain rule
$$p(y_1\dots y_n|{\bf x}) =  \prod_{t=1}^{n} p(y_t|y_1\dots y_{t-1},{\bf x})$$
resulting in decoding each word depending on all previous decoding results, thus hindering the parallelizability.
In the SAT, we extend the standard word-level chain rule to the group-level chain rule.

We first divide the word sequence $y_1\dots y_n$ into consecutive groups
\begin{align*}
&G_1,G_2,\dots ,G_{[(n-1)/K]+1} =\\
&y_1\dots y_K, y_{K+1}\dots y_{2K}, \dots, y_{[(n-1)/K]\times K+1}\dots y_{n}
\end{align*}
where $[\cdot]$ denotes floor operation, $K$ is the group size, and also the indicator of parallelizability. The larger the $K$, the higher the parallelizability. Except for the last group, all groups must contain $K$ words.
Then comes the group-level chain rule 
$$p(y_1\dots y_n|{\bf x}) =  \prod_{t=1}^{[(n-1)/K]+1} p(G_t|G_1\dots G_{t-1},{\bf x})$$

This group-level chain rule avoids the dependencies between consecutive words if they are in the same group. With group-level chain rule, the model no longer produce words one by one as the Transformer, but rather group by group.
In next subsections, we will show how to implement the model in detail.

\subsection{Long-Distance Prediction}
\begin{figure}
	\centering
	\includegraphics[width=\linewidth]{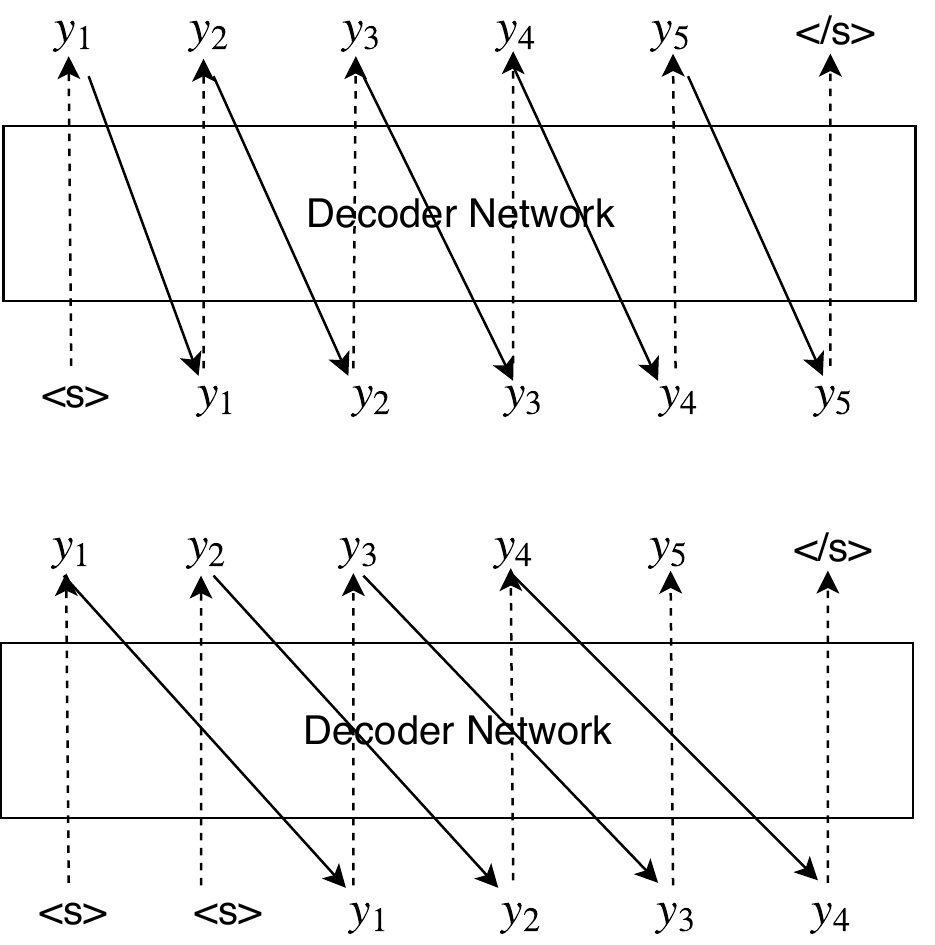}
	\caption{\label{fig:inputs}Short-distance prediction (top) and long-distance prediction (bottom).}
\end{figure}

In autoregressive models, to predict $y_t$, the model should be fed with the previous word $y_{t-1}$. We refer it as \emph{short-distance prediction}.
In the SAT, however, we feed $y_{t-K}$ to predict $y_t$, to which we refer as \emph{long-distance prediction}.
At the beginning of decoding, we feed the model with $K$ special symbols ${<}\text{s}{>}$ to predict $y_1\dots y_K$ in parallel. 
Then $y_1\dots y_K$ are fed to the model to predict $y_{K+1}\dots y_{2K}$ in parallel. This process will continue until a terminator ${<}{/}\text{s}{>}$ is generated.
Figure \ref{fig:inputs} gives illustrations for both short and long-distance prediction.

\subsection{Relaxed Causal Mask}
In the Transformer decoder, the causal mask is a lower triangular matrix, which strictly prevents earlier decoding steps from peeping information from later steps. We denote it as \emph{strict causal mask}.
However, in the SAT decoder, strict causal mask is not a good choice. 
As described in the previous subsection, in long-distance prediction, the model predicts $y_{K+1}$ by feeding with $y_1$. With strict causal mask, the model can only access to $y_1$ when predict $y_{K+1}$, which is not reasonable since $y_1\dots y_K$ are already produced. It is better to allow the model to access to $y_1\dots y_K$ rather than only $y_1$ when predict $y_{K+1}$.

Therefore, we use a coarse-grained lower triangular matrix as the causal mask that allows peeping later information in the same group. We refer to it as \emph{relaxed causal mask}. Given the target length $n$ and the group size $K$, relaxed causal mask $M \in \mathbb{R}^{n\times n}$ and its elements are defined below:
$$ M[i][j]=
\begin{cases}
1 & \text{if $j < ([(i-1)/K]+1)\times K$} \\
0 & \text{other}
\end{cases}$$

For a more intuitive understanding, Figure \ref{fig:masks} gives a comparison between strict and relaxed causal mask.

\begin{figure}
	\centering
	$\left[\begin{matrix}
	1&\boldsymbol{0}&0&0&0&0\\
	1&1&0&0&0&0\\
	1&1&1&\boldsymbol{0}&0&0\\
	1&1&1&1&0&0\\
	1&1&1&1&1&\boldsymbol{0}\\
	1&1&1&1&1&1
	\end{matrix}\right]$\quad
	$\left[\begin{matrix}
	1&\boldsymbol{1}&0&0&0&0\\
	1&1&0&0&0&0\\
	1&1&1&\boldsymbol{1}&0&0\\
	1&1&1&1&0&0\\
	1&1&1&1&1&\boldsymbol{1}\\
	1&1&1&1&1&1
	\end{matrix}\right]$
	\caption{\label{fig:masks}Strict causal mask (left) and relaxed causal mask (right) when the target length $n=6$ and the group size $K=2$.
	We mark their differences in bold.}
\end{figure}

\subsection{The SAT}
\label{sec:sat}
Using group-level chain rule instead of word-level chain rule, long-distance prediction instead of short-distance prediction, and relaxed causal mask instead of strict causal mask, we successfully extended the Transformer to the SAT. 
The Transformer can be viewed as a special case of the SAT, when the group size $K$ = 1.
The non-autoregressive Transformer (NAT) described in \newcite{gu2017non} can also be viewed as a special case of the SAT, when the group size $K$ is not less than maximum target length. 

Table \ref{tab:complexity} gives the theoretical complexity and acceleration of the model. We list two search strategies separately: beam search and greedy search.
Beam search is the most prevailing search strategy. However, it requires the decoder states to be updated once every word is generated, thus hinders the decoding parallelizability. When decode with greedy search, there is no such concern, therefore the parallelizability of the SAT can be maximized.

\begin{table}
	\resizebox{\linewidth}{!}{
	\begin{tabular}{l|c|c}
		\hline
		Model & Complexity	&	Acceleration\\
		\hline
		Transformer & $N(a+b)$ &  1\\
		\hline
		SAT (beam search) & $\frac{N}{K}a + Nb$ &  $K(\frac{a+b}{a+Kb})$\\
		\hline
		SAT (greedy search) & $\frac{N}{K}(a+b)$ & $K$ \\
		\hline
	\end{tabular}}
	\caption{\label{tab:complexity}Theoretical complexity and acceleration of the SAT. $a$ denotes the time consumed on the decoder network (calculating a distribution over the target vocabulary) each time step and $b$ denotes the time consumed on search (searching for top scores, expanding nodes and pruning). In practice, $a$ is usually much larger than $b$ since the network is deep.}
\end{table}

\section{Experiments}
We evaluate the proposed SAT on English-German and Chinese-English translation tasks.

\subsection{Experimental Settings}
\noindent{\bf Datasets} \quad
For English-German translation, we choose the corpora provided by WMT 2014 \cite{bojar2014findings}. We use the newstest2013 dataset for development, and the newstest2014 dataset for test.
For Chinese-English translation, the corpora we use is extracted from LDC\footnote{The corpora include LDC2002E18, LDC2003E14, LDC2004T08 and LDC2005T0.}. We chose the NIST02 dataset for development, and the NIST03, NIST04 and NIST05 datasets for test.
For English and German, we tokenized and segmented them into subword symbols using byte-pair encoding (BPE) \cite{sennrich2015neural} to restrict the vocabulary size. As for Chinese, we segmented sentences into characters.
For English-German translation, we use a shared source and target vocabulary.
Table \ref{tab:corpora} summaries the two corpora.
\begin{table}
	\begin{tabular}{c|c|c|c}
	\hline
	& \multirow{2}{*}{Sentence Number}	&	\multicolumn{2}{c}{Vocab Size}\\
	\cline{3-4}
	&	& Source & Target \\
	\hline
	EN-DE & 4.5M & 36K & 36K \\
	\hline
	ZH-EN & 1.8M & 9K & 34K \\
	\hline
	\end{tabular}
	\caption{\label{tab:corpora}Summary of the two corpora.}
\end{table}

\begin{table*}[!htb]
	\centering
	\resizebox{0.8\linewidth}{!}{
		\begin{tabular}{l|c|c|c|c|c}
			\hline
			Model	& Beam Size	& BLEU	&	Degeneration	&	Latency	&	Speedup\\
			\hline
			\multirow{2}{*}{Transformer}	&	4	&	27.11 & 0\%	& 346ms	&	1.00$\times$\\
			&   1    &   26.01 & 4\% & 283ms &   1.22$\times$\\
			\hline
			\multirow{2}{*}{Transformer, $N$=2}	&	4	&	24.30 & 10\%	& 163ms	&	2.12$\times$\\
			&   1    &   23.37 & 14\% & 113ms &   3.06$\times$\\
			\hline
			NAT \cite{gu2017non}	&	- & 17.69	&	25\% & 39ms &	15.6$\times$\\
			NAT (rescroing 10)	& - & 18.66 &	20\%& 79ms &	7.68$\times$\\
			NAT (rescroing 100)	& - & 19.17 &	18\% & 257ms &	2.36$\times$\\
			\hline
			LT \cite{kaiser2018fast} & - & 19.80 &	27\%	& 105ms & -\\
			LT (rescoring 10) & - & 21.00 &	23\%	& -	&	- \\
			LT (rescoring 100) & - & 22.50 &	18\%	& - &	- \\
			\hline
			IRNAT \cite{lee2018deterministic} & - & 18.91 & 22\% & - & 1.98$\times$ \\
			\hline
			\multicolumn{6}{c}{\emph{This Work}} \\
			\hline
			\multirow{2}{*}{SAT, $K$=2}	& 4	&	26.90	&	1\% & 229ms& 1.51$\times$ \\
			& 1 &   26.09	  &	  4\%&	167ms & 2.07$\times$\\
			\hline
			\multirow{2}{*}{SAT, $K$=4}	& 4 &	25.71	&	5\% &149ms	& 2.32$\times$\\
			& 1 &   24.67	  &	  9\%&	91ms & 3.80$\times$\\
			\hline
			\multirow{2}{*}{SAT, $K$=6}	& 4 &	24.83	&	8\% &116ms	& 2.98$\times$\\
			& 1 &	23.93	&	12\% &62ms	& 5.58$\times$ \\
			\hline
		\end{tabular}}
	\caption{\label{tab:enderes}Results on English-German translation. Latency is calculated on a single NVIDIA TITAN Xp without batching. For comparison, we also list results reported by \newcite{gu2017non,kaiser2018fast,lee2018deterministic}. Note that \newcite{gu2017non,lee2018deterministic} used PyTorch as their platform, but we and \newcite{kaiser2018fast} used TensorFlow. Even on the same platform, implementation and hardware may not exactly be the same. Therefore, it is not fair to directly compare BLEU and latency. A fairer way is to compare performance degradation and speedup, which are calculated based on their own baseline.}
\end{table*}

\vspace{5pt}
\noindent{\bf Baseline} \quad
We use the base Transformer model described in \newcite{vaswani2017attention} as the baseline, where $d_{model}=512 \text{ and } N=6$.
In addition, for comparison, we also prepared a lighter Transformer model, in which two encoder/decoder blocks are used ($N$ = 2), and other hyper-parameters remain the same.

\vspace{5pt}
\noindent{\bf Hyperparameters} \quad
Unless otherwise specified, all hyperparameters are inherited from the base Transformer model. 
We try three different settings of the group size $K$: $K$ = 2, $K$ = 4, and $K$ = 6.
For English-German translation, we share the same weight matrix between the source and target embedding layers and the pre-softmax linear layer.
For Chinese-English translation, we only share weights of the target embedding layer and the pre-softmax linear layer.

\vspace{5pt}
\noindent{\bf Search Strategies} \quad
We use two search strategies: beam search and greedy search. As mentioned in Section \ref{sec:sat}, these two strategies lead to different parallelizability. When beam size is set to 1, greedy search is used, otherwise, beam search is used.

\vspace{5pt}
\noindent{\bf Knowledge Distillation} \quad
Knowledge distillation \cite{hinton2015distilling,kim2016sequence} describes a class of methods for training a smaller \emph{student} network to perform better by learning from a larger \emph{teacher} network. For NMT, \newcite{kim2016sequence} proposed a sequence-level knowledge distillation method. 
In this work, we apply this method to train the SAT using a pre-trained autoregressive Transformer network.
This method consists of three steps: (1) train an autoregressive Transformer network (the \emph{teacher}), (2) run beam search over the training set with this model and (3) train the SAT (the \emph{student}) on this new created corpus.

\vspace{5pt}
\noindent{\bf Initialization} \quad
Since the SAT and the Transformer have only slight differences in their architecture (see Figure \ref{fig:transformer}), in order to accelerate convergence, we use a pre-trained Transformer model to initialize some parameters in the SAT. These parameters include all parameters in the encoder, source and target word embeddings, and pre-softmax weights. Other parameters are initialized randomly.
In addition to accelerating convergence, we find this method also slightly improves the translation quality.

\vspace{5pt}
\noindent{\bf Training} \quad
Same as \newcite{vaswani2017attention}, we train the SAT by minimize cross-entropy with label smoothing.
The optimizer we use is Adam \cite{kingma2015adam} with $\beta_1=0.9$, $\beta_2=0.98$ and $\varepsilon=10^{-9}$. We change the learning rate during training using the learning rate funtion described in \newcite{vaswani2017attention}.
All models are trained for 10K steps on 8 NVIDIA TITAN Xp with each minibatch consisting of about 30k tokens.
For evaluation, we average last five checkpoints saved with an interval of 1000 training steps.

\vspace{5pt}
\noindent{\bf Evaluation Metrics} \quad
We evaluate the translation quality of the model using BLEU score \cite{papineni2002bleu}.

\vspace{5pt}
\noindent{\bf Implementation} \quad
We implement the proposed SAT with \emph{TensorFlow} \cite{abadi2016tensorflow}. The code and resources needed for reproducing the results are released at \url{https://github.com/chqiwang/sa-nmt}.

\subsection{Results on English-German}
Table \ref{tab:enderes} summaries results of English-German translation.
According to the results, the translation quality of the SAT gradually decreases as $K$ increases, which is consistent with intuition.
When $K$ = 2, the SAT decodes 1.51$\times$ faster than the Transformer and is almost lossless in translation quality (only drops 0.21 BLEU score).
With $K$ = 6, the SAT can achieve 2.98$\times$ speedup while the performance degeneration is only 8\%.

When using greedy search, the acceleration becomes much more significant.
When $K$ = 6, the decoding speed of the SAT can reach about $5.58\times$ of the Transformer while maintaining 88\% of translation quality. Comparing with \newcite{gu2017non,kaiser2018fast,lee2018deterministic}, the SAT achieves a better balance between translation quality and decoding speed.
Compared to the lighter Transformer ($N$ = 2), with $K$ = 4, the SAT achieves a higher speedup with significantly better translation quality.

In a real production environment, it is often not to decode sentences one by one, but batch by batch. To investigate whether the SAT can accelerate decoding when decoding in batches, we test the decoding latency under different batch size settings.  As shown in Table \ref{tab:latency}, the SAT significantly accelerates decoding even with a large batch size.
\begin{table}
	\centering
	\begin{tabular}{l|c|c|c|c}
		\hline
		Model	&	b=1	& b=16	& b=32	& b=64 \\
		\hline
		Transformer  & 346ms  & 58ms & 53ms & 56ms\\
		\hline
		SAT, $K$=2	& 229ms  & 38ms & 32ms & 32ms\\
		SAT, $K$=4	& 149ms   & 24ms & 21ms &20ms\\
		SAT, $K$=6  & 116ms   & 20ms & 17ms & 16ms\\
		\hline
	\end{tabular}
	\caption{\label{tab:latency}Time needed to decode one sentence under various batch size settings. A single NVIDIA TIAN Xp is used in this test.}
\end{table}

It is also good to know if the SAT can still accelerate decoding on CPU device that does not support parallel execution as well as GPU.
Results in Table \ref{tab:cpu_latency} show that even on CPU device, the SAT can still accelerate decoding significantly.
\begin{table}
	\centering
	\begin{tabular}{l|c|c|c|c}
		\hline
		Model	&	$K$=1	& $K$=2	& $K$=4	& $K$=6 \\
		\hline
		Latency  & 1384ms  & 607ms & 502ms & 372ms\\
		\hline
	\end{tabular}
	\caption{\label{tab:cpu_latency}Time needed to decode one sentence on CPU device. Sentences are decoded one by one without batching. $K$=1 denotes the Transformer.}
\end{table}

\begin{table*}[!htb]
\centering
\resizebox{\linewidth}{!}{
	\begin{tabular}{l|c|c|c|c|c|c|c|c}
		\hline
		\multirow{2}{*}{Model}	& \multirow{2}{*}{Beam Size}	&	\multicolumn{4}{c|}{BLEU} & \multirow{2}{*}{Degeneration} &	\multirow{2}{*}{Lattency}&	\multirow{2}{*}{Speedup}\\
		\cline{3-6}
		& &	NIST03	& NIST04	&	NIST05	&	Averaged	&& \\
		\hline
		\multirow{2}{*}{Transformer}	&  4 &	40.74 & 40.54 & 40.48 & 40.59 & 0\% & 410ms & 1.00$\times$\\
		& 1 & 39.56 & 39.72 & 39.61 & 39.63 & 2\% & 302ms & 1.36$\times$\\
		\hline
		\multirow{2}{*}{Transformer, $N$=2}	& 4 & 37.30	&	38.55	&	36.87	&	37.57	&	7\% & 169ms & 2.43$\times$\\
		& 1 & 36.26 & 37.19 & 35.50 & 36.32 & 11\% & 117ms & 3.50$\times$\\
		\hline
		\multicolumn{9}{c}{\emph{This Work}} \\
		\hline
		\multirow{2}{*}{SAT, $K$=2}	& 4 &	39.13 & 40.04 & 39.55 & 39.57 &	3\% & 243ms & 1.69$\times$\\
		& 1 & 37.94 & 38.73 & 38.43 & 38.37 & 5\% & 176ms & 2.33$\times$\\
		\hline
		\multirow{2}{*}{SAT, $K$=4}	& 4 &	37.08	&	38.06	&	37.12	&	37.42	&	8\% & 152ms& 2.70$\times$\\
		& 1 & 35.77 & 36.43 & 35.04 & 35.75 & 12\% &94ms & 4.36$\times$\\
		\hline
		\multirow{2}{*}{SAT, $K$=6}	& 4 & 34.61	&	36.29	&	35.06	&	35.32 &	13\% & 129ms & 3.18$\times$\\
		& 1	&33.44	&34.54	&33.28	&33.75 & 17\% &64ms & 6.41$\times$\\
		\hline
	\end{tabular}}
	\caption{\label{tab:zhenres}Results on Chinese-English translation. Latency is calculated on NIST02.}
\end{table*}

\subsection{Results on Chinese-English}
Table \ref{tab:zhenres} summaries results on Chinese-English translation. With $K$ = 2, the SAT decodes 1.69$\times$ while maintaining 97\% of the translation quality. In an extreme setting where $K$ = 6 and beam size = 1, the SAT can achieve 6.41$\times$ speedup while maintaining 83\% of the translation quality.

\subsection{Analysis}
\begin{figure}
	\centering
	\includegraphics[width=\linewidth]{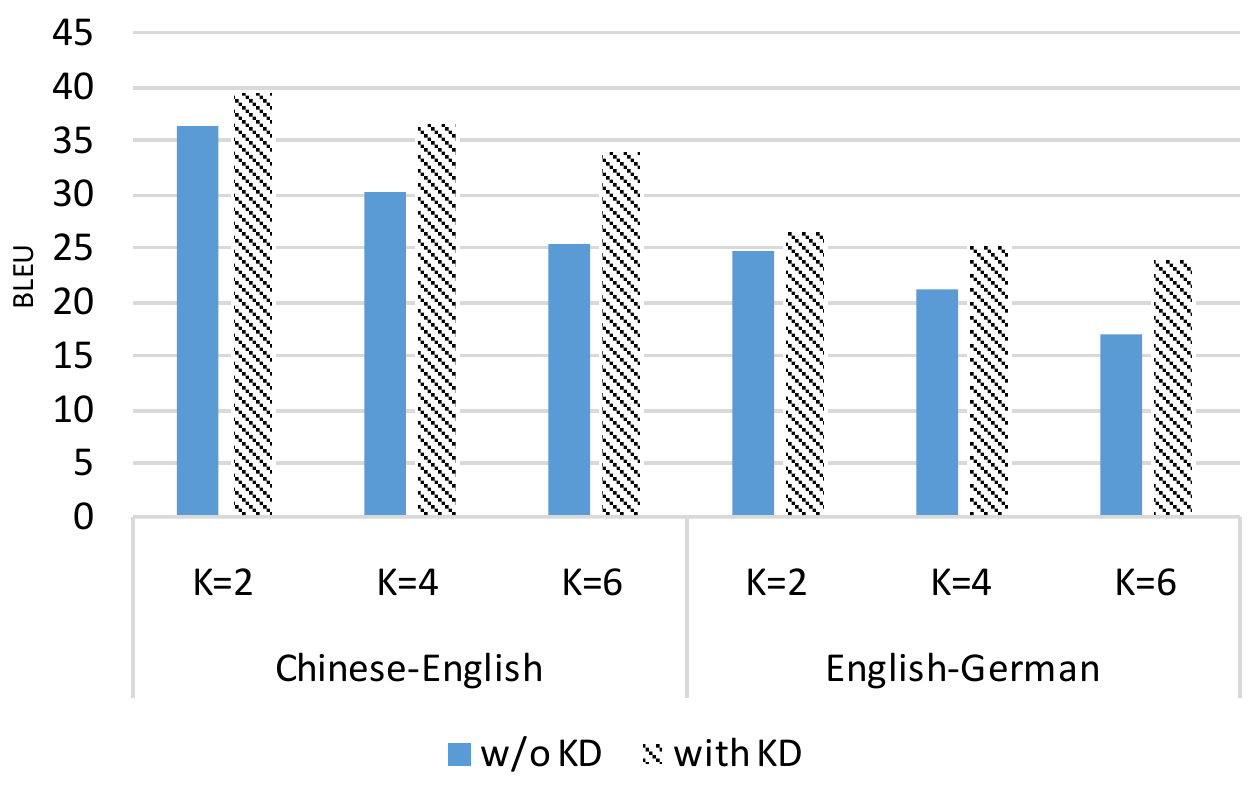}
	\caption{\label{fig:kd}Performance of the SAT with and without sequence-level knowledge distillation.}
\end{figure}

\vspace{5pt}
\noindent{\bf Effects of Knowledge Distillation} \quad
As shown in Figure \ref{fig:kd}, sequence-level knowledge distillation is very effective for training the SAT. For larger  $K$, the effect is more significant.
This phenomenon is echoing with observations by \newcite{gu2017non,oord2017parallel,lee2018deterministic}.
In addition, we tried word-level knowledge distillation \cite{kim2016sequence} but only a slight improvement was observed.

\begin{figure}
	\centering
	\includegraphics[width=\linewidth]{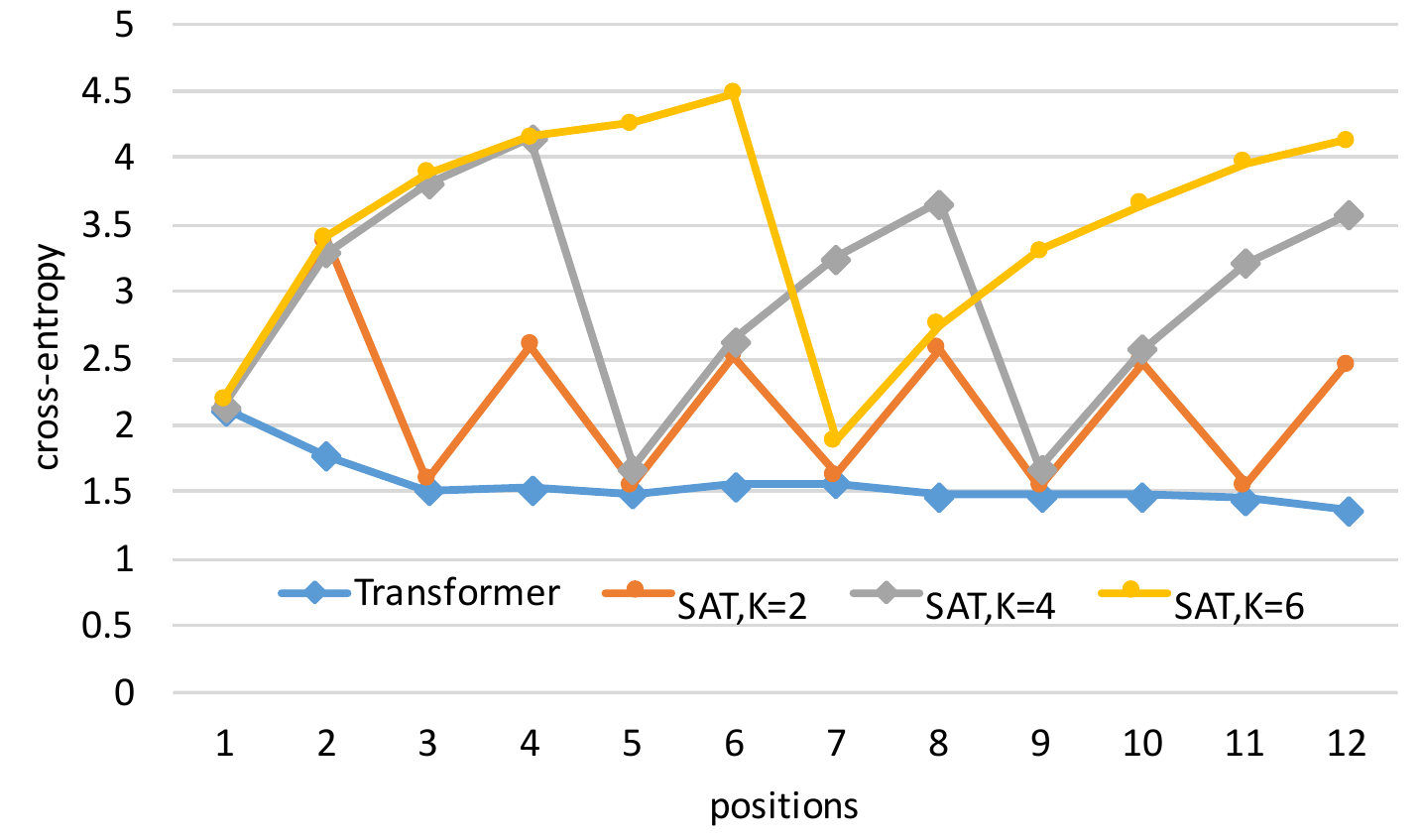}
	\caption{\label{fig:loss}Position-wise cross-entropy for various models on English-German translation.}
\end{figure}

\vspace{5pt}
\noindent{\bf Position-Wise Cross-Entropy} \quad
In Figure \ref{fig:loss}, we plot position-wise cross-entropy for various models. To compare with the baseline model, the results in the figure are from models trained on the original corpora, i.e., without knowledge distillation.
As shown in the figure, position-wise cross-entropy has an apparent periodicity with a period of $K$. For positions in the same group, the position-wise cross-entropy increase monotonously, which indicates that the long-distance dependencies are always more difficult to model than short ones. 
It suggests the key to further improve the SAT is to improve the ability of modeling long-distance dependencies.

\begin{table*}[!htb]
	\centering
	\includegraphics[width=\linewidth]{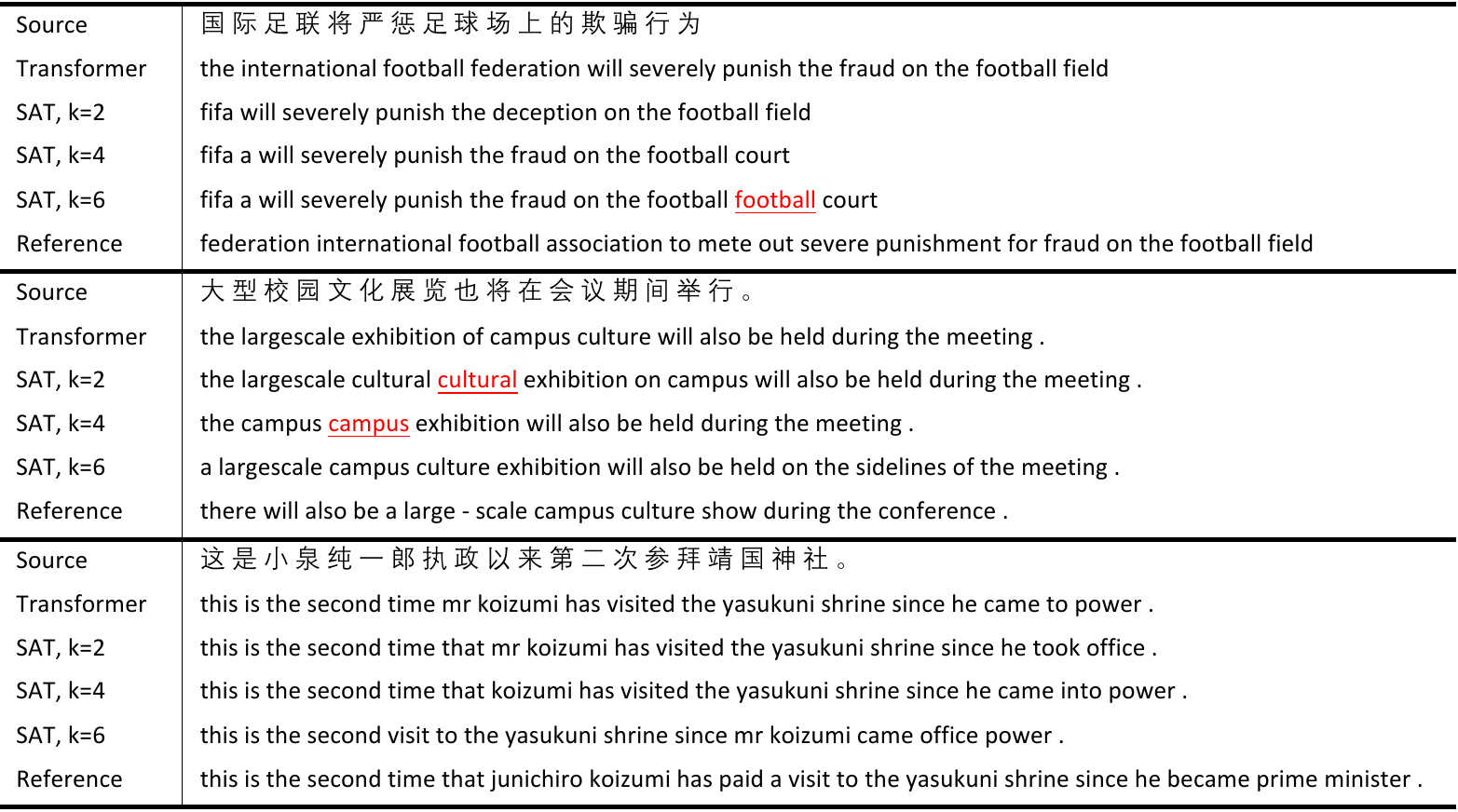}
	\caption{\label{tab:cases}Three sample Chinese-English translations by the SAT and the Transformer. We mark repeated words or phrases by red font and underline.}
\end{table*}

\vspace{5pt}
\noindent{\bf Case Study} \quad
Table \ref{tab:cases} lists three sample Chinese-English translations from the development set. As shown in the table, even when produces $K$ = 6 words at each time step, the model can still generate fluent sentences.
As reported by \newcite{gu2017non}, instances of repeated words or phrases are most prevalent in their non-autoregressive model. In the SAT, this is also the case. This suggests that we may be able to improve the translation quality of the SAT by reducing the similarity of the output distribution of adjacent positions.

\section{Conclusion}
In this work, we have introduced a novel model for faster sequence generation based on the Transformer \cite{vaswani2017attention}, which we refer to as the semi-autoregressive Transformer (SAT).
Combining the original Transformer with group-level chain rule, long-distance prediction and relaxed causal mask, the SAT can produce multiple consecutive words at each time step, thus speedup decoding significantly.
We conducted experiments on English-German and Chinese-English translation. Compared with previously proposed non-autoregressive models \cite{gu2017non,lee2018deterministic,kaiser2018fast}, the SAT achieves a better balance between translation quality and decoding speed. On WMT'14 English-German translation, the SAT achieves 5.58$\times$ speedup while maintaining 88\% translation quality, significantly better than previous methods. When produces two words at each time step, the SAT is almost lossless (only 1\% degeneration in BLEU score).

In the future, we plan to investigate better methods for training the SAT to further shrink the performance gap between the SAT and the Transformer.
Specifically, we believe that the following two directions are worth study.
First, use object function beyond maximum likelihood to improve the modeling of long-distance dependencies.
Second, explore new method for knowledge distillation.
We also plan to extend the SAT to allow the use of different group sizes $K$ at different positions, instead of using a fixed value.

\section*{Acknowledgments}
We would like to thank the anonymous reviewers for their valuable comments. We also thank Wenfu Wang, Hao Wang for helpful discussion and Linhao Dong, Jinghao Niu for their help in paper writting.

\bibliography{sanmt}
\bibliographystyle{acl_natbib_nourl}

\appendix

\end{document}